\pdfoutput=1

\documentclass[11pt]{article}

\usepackage[]{acl}
\usepackage{times}
\usepackage{latexsym}

\usepackage[T1]{fontenc}

\usepackage[utf8]{inputenc}

\usepackage{microtype}

\usepackage{graphicx}
\usepackage{subcaption}
\usepackage{amsmath}
\usepackage{amsfonts}
\usepackage{mathrsfs}
\usepackage{multirow}
\usepackage{booktabs}
\usepackage{amssymb}
\usepackage{pifont}

\newcommand{\knowexpert}{\texttt{KnowExpert}}

\title{Retrieval-Free Knowledge-Grounded Dialogue Response Generation with Adapters}

\author{Yan Xu\thanks{$^*$ These two authors contributed equally.}, Etsuko Ishii$^*$, Samuel Cahyawijaya, Zihan Liu, Genta Indra Winata, \\ {\bf Andrea Madotto, Dan Su, Pascale Fung} \\
Center for Artificial Intelligence Research (CAiRE)\\
The Hong Kong University of Science and Technology, Clear Water Bay, Hong Kong\\
\tt \{yxucb,eishii\}@connect.ust.hk, pascale@ece.ust.hk
}

\begin{document}
\maketitle
\begin{abstract}
To diversify and enrich generated dialogue responses, knowledge-grounded dialogue has been investigated in recent years. 
The existing methods tackle the knowledge grounding challenge by retrieving the relevant sentences over a large corpus and augmenting the dialogues with explicit extra information. Despite their success, however, the existing works have drawbacks in inference efficiency. This paper proposes \knowexpert, a framework to bypass the explicit retrieval process and inject knowledge into the pre-trained language models with lightweight adapters and adapt to the knowledge-grounded dialogue task. To the best of our knowledge, this is the first attempt to tackle this challenge without retrieval in this task under an open-domain chit-chat scenario. The experimental results show that \knowexpert~performs comparably with some retrieval-based baselines while being time-efficient in inference, demonstrating the effectiveness of our proposed method.\footnote{Our code and models are available at \url{https://github.com/HLTCHKUST/KnowExpert}.}
\end{abstract}

\section{Introduction}

Numerous studies in recent years have established sophisticated techniques to build open-domain dialogue systems. Although such systems can generate fluent and grammatically correct responses based on the dialogue history, they are unsatisfactory compared to human-to-human conversations. 
One primary reason is that existing dialogue systems are incapable of understanding and leveraging relevant knowledge, resulting in superficial and un-intelligent responses when they dive into a specific topic~\cite{li2020zero-resource}. To overcome this limitation, many research works have focused on developing knowledge-grounded dialogue (KGD) systems~\citep{dinan2018wizard,chen2020bridging,zhao2020knowledge}.




\begin{figure*}[!t]
    \centering
    \begin{minipage}{.64\textwidth}
      \centering
      \includegraphics[width=\linewidth]{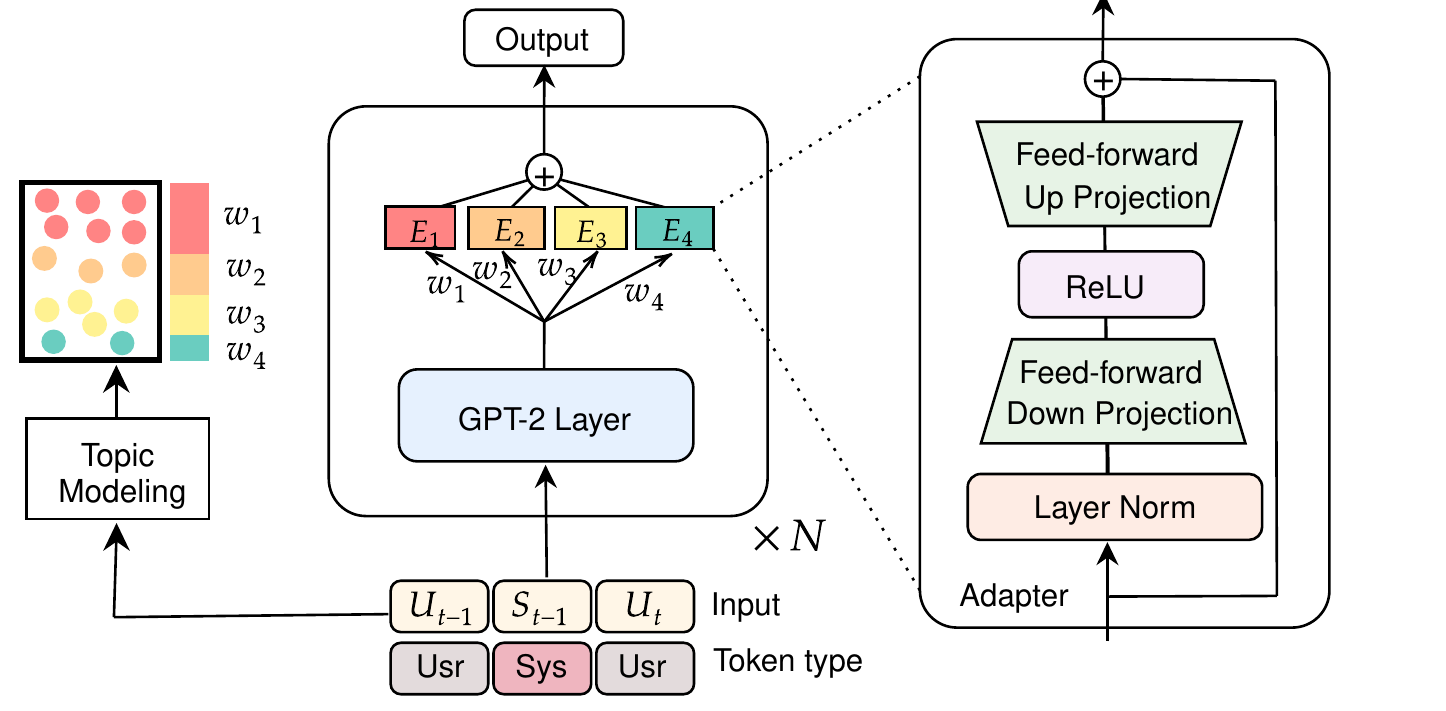}
      \caption{High-level architecture of the model. Taking a dialogue history as an input, the adapters are inserted upon the GPT-2 layers, acting as the knowledge experts, to enhance the response generation with the help from a topic model which assigns weights over the knowledge experts.}
      \label{fig:architecture}
    \end{minipage}%
    \hfill
    \begin{minipage}{.34\textwidth}
      \centering
      \includegraphics[width=\linewidth]{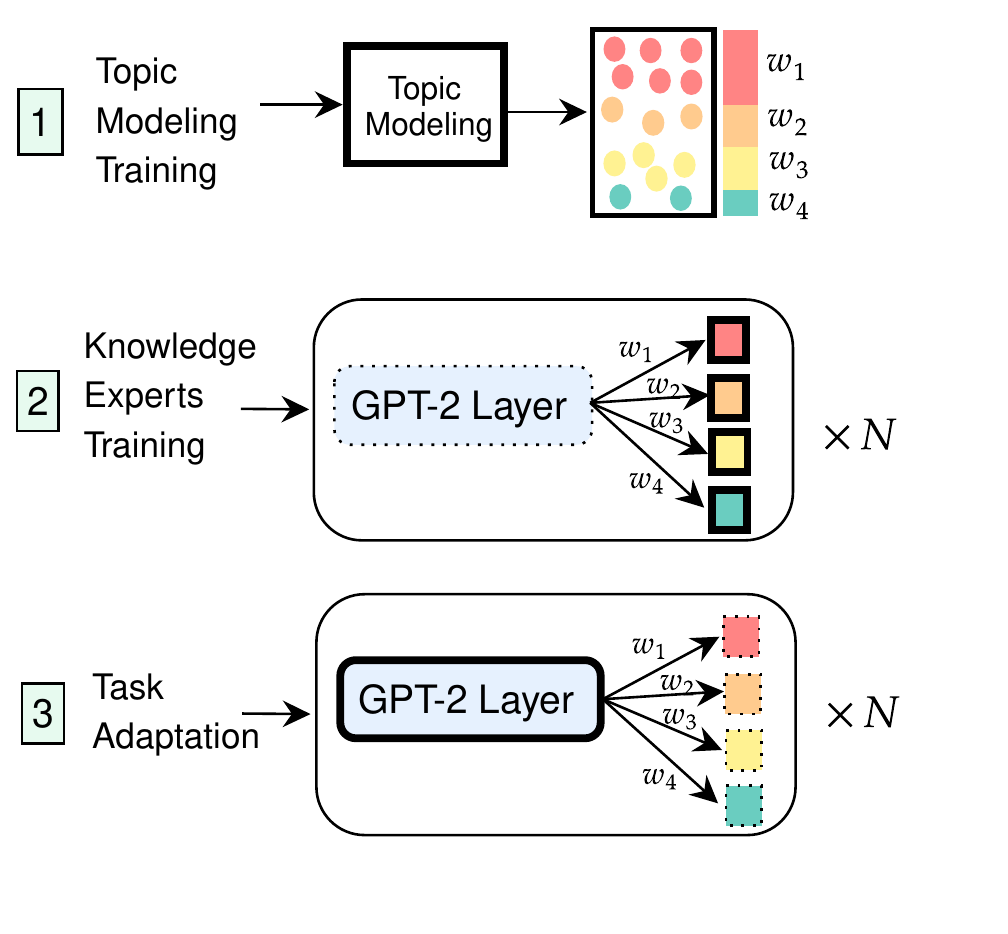}
      \caption{Illustration of the training procedure, where the thick lined modules are trained while the rest (dash lined) kept frozen in each training step.}
      \label{fig:pipeline}
    \end{minipage}
\end{figure*}

To equip the ability to incorporate knowledge, many recently proposed KGD systems~\citep{lian2019learning,kim2019sequential,roller2020recipes,chen2020bridging,zhao2020knowledge} comprise the following modules: (1) Knowledge Retrieval, for retrieving the related knowledge sentences from a large corpus (e.g., Wikipedia); (2) Knowledge Selection, for selecting the most relevant knowledge sentences for generation; and (3) Knowledge-augmented Generation, for augmenting the retrieved knowledge and conversation history to generate more knowledgeable responses.
The key to this approach is the explicit retrieval phase to enhance the quality of generated responses.

Despite demonstrating remarkable progress and promising performance on the KGD task, the retrieval-based approaches have drawbacks in their efficiency. First, knowledge retrieval in corpora requires a model to search over a large amount of data, consuming considerable memory resources to store the whole knowledge corpus. It also takes additional processing time to retrieve knowledge and conduct further knowledge selection. Second, adding knowledge as additional context to the language generation model also causes significant computation overhead, which slows the language generation process. Efficiency plays an essential role in the practical use of dialogue systems, and it is necessary to limit resource requirements so as to generate responses faster and support more active users.

Recently, large pre-trained language models (LMs) have been shown to have the capability to carry implicit knowledge~\cite{wang2020k,lauscher2020common}, which can be further applied to downstream classification tasks~\cite{shwartz2020unsupervised}. Many existing works have proved that the ``knowledge'' can be embedded in the pre-training process~\citep{brown2020language}. The explorations on the closed-book question answering (QA) task~\cite{petroni2019language,roberts2020much,wang2021can} with large pre-trained LMs also indicates the potential of leveraging the knowledge embedded inside LMs. For task-oriented dialogue systems, \citet{madotto2020learning} store knowledge bases (KBs) of different sizes directly into the model parameters by aligning auto-extracted dialogue templates with the corresponding KBs for each data sample. Based on their success in other tasks, LMs have potential to apply their implicit knowledge for open-domain KGD tasks. However, our scenario is different from both the closed-book QA and task-oriented dialogue tasks, where given a question or user query, relevant knowledge choices are highly constrained by the inputs. In contrast, open-domain chit-chat suffers much from the one-to-many issue between the inputs and possible outputs. In other words, given the inputs on a specific topic, the choice of knowledge candidates is varying, which brings new challenges to embedding knowledge in this task.


Inspired by the previous explorations on other tasks, we propose to tackle the KGD challenge by using the implicit knowledge in LMs under the open-domain chit-chat scenario. 
In contrast to existing KGD systems, we bypass the retrieval step and propose a framework, \knowexpert, to learn the knowledge corpus with the parameters of pre-trained LMs and incorporate the acquired knowledge for KGD generation. In the model, lightweight adapters~\cite{bapna2019simple} are inserted into the pre-trained GPT-2~\cite{radford2019language}, acting as knowledge experts. Taking advantage of latent topics, the knowledge sentences are embedded into different knowledge experts by pseudo-conversation style training, while the latent topics measure the relevance between the dialogue samples and the clusters. 
We thus fine-tune LM layers where frozen pre-trained adapters are inserted for task adaptation. 
Experimental results show that \knowexpert performs comparably with some strong retrieval-based baselines, while its inference process is much more efficient since extra knowledge sentences are not required as a component of the inputs. 

Our contributions are three-fold: (1) to the best of our knowledge, we are the first to explore learning prior knowledge with generative models for KGD tasks under open-domain chit-chat scenario; (2) our model bypass an explicit knowledge retrieval process, and has constant inference time regardless of the size of the knowledge corpus; and (3) our model performs comparably with some strong baselines and shows that a purely generation-based method for the KGD task is promising.

\begin{figure*}[t]
  \centering
  \includegraphics[width=\linewidth]{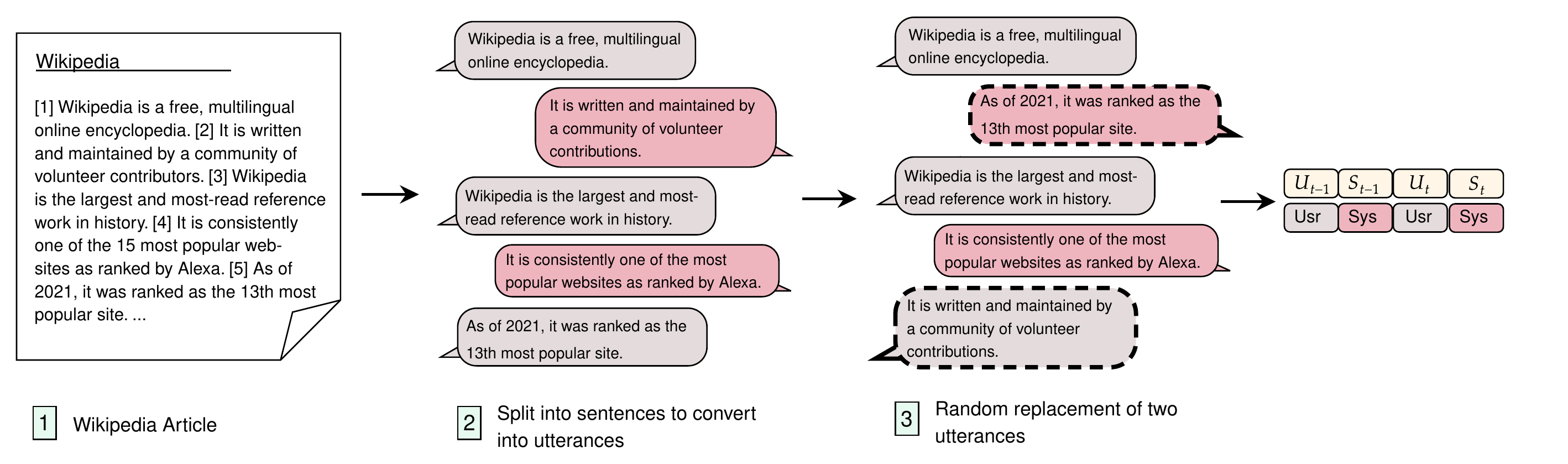} 
  \caption{The demonstration of the procedure of converting a document in the knowledge corpus (e.g., a Wikipedia article) into the pseudo-conversation style. First, the article is split into sentences so that each to represent one utterance. Then, random two utterances are permuted to avoid over-fitting (presented with dashed lines).}
  \label{fig:pseudo}
\end{figure*}

\section{Related Work}
\subsection{Knowledge-Grounded Dialogue}


The KGD task requires models to identify relevant knowledge sentences from a large corpus for grounding the response generation.
Information retrieval (IR) systems, such as TF-IDF, can quickly retrieve related knowledge over a large corpus. However, the effectiveness is limited as they can only be leveraged to coarsely retrieve several relevant documents. However, providing the models with more documents may not improve the system since it will bring more noise into the inputs. What is more, the length of the packed inputs could exceed the length limitation of the LMs. Thus, the existing works still conduct further fine-grained knowledge selection to improve the accuracy of the knowledge retrieval process, which is one of the critical problems in the KGD task. Motivated by this, latent variables have been introduced to minimize the gap between prior and the posterior knowledge selection~\cite{zhao2019low,kim2019sequential,lian2019learning,chen2020bridging}. \citet{zhao2020knowledge} explores the strategy to better rank the knowledge sentences, avoiding the most relevant candidates becoming truncated in the input sequences. Some existing KGD systems generate the knowledge first for further response generation. \cite{zhou2021think} train the model to generate implicit knowledge sentences for open-domain dialogue response generation. Instead of training the large pre-trained LMs, \cite{liu2022multi} leverage prompts for knowledge and response generation.  
\citet{cui2021knowledge} proposes the knowledge-enhanced fine-tuning for better handling the unseen entities in the conversation history. They also evaluate the model when there is no knowledge sentence as inputs during inference. However their proposed method only focus on the problem of unseen entities, whereas it is less helpful on the seen domain. In this paper, we propose a new promising direction to bypass the retrieval step and better leverage power of the pretrained LMs for knowledge-grounded diaogue generation.

\subsection{Knowledge Retrieval in LMs}
The concept of \textit{knowledge retrieval in LMs} started with the proposal of the LAMA benchmark~\cite{petroni2019language}, which heavily relies on prompts. By constructing the prompts as “fill-in-the-blank” cloze statements, pre-trained LMs can predict the factual knowledge~\cite{petroni2019language,shin2020autoprompt}.
The application of the idea of knowledge retrieval in LMs also appears in closed-book QA tasks. \citet{roberts2020much} investigates a simple fine-tuning technique on multiple QA datasets and proves that T5~\cite{raffel2019exploring} can pack Wikipedia knowledge into its parameters. 



\subsection{Inference Efficiency in Language Model}

Recent progress in natural language processing, including dialogue systems, has been benefited by Transformer-based large pre-trained LMs, yet current ``best performing" models tend to have a more complex architecture with more parameters, which is not ideal considering inference in practical application. Many modified Transformer architectures have been explored to speed up inference latency while maintaining performance, for example, by leveraging knowledge distillation to compress or reduce the number of parameters~\citep{tang2019distilling, sanh2019distilbert, jiao2020tinybert, sun2020mobilebert}, by performing a simple decomposition in lower layers~\citep{cao2020deformer}, or by converting a structured decoder into a non-autoregressive module~\citep{sun2019fast}. Contrasting previous works, we emphasize the inference efficiency of our proposed framework in shortening the input sequences by removing the external knowledge components and reducing the storage resources needed, and we provide a faster inference process when scaling up the knowledge corpus.

\section{Methodology}
In this section, we present the framework and learning algorithm of \knowexpert. First, we offer several preliminary definitions used throughout the paper. Second, we explain the architecture of \knowexpert. Finally, we describe the strategy to train the framework.

\subsection{Preliminary Definition}
\label{sec:notation}
We denote a dialogue dataset as $\{\mathcal{D}^n\}_{n=1}^N$, and the dialogue history at turn $t$ as $\mathcal{D}_{t} = \{ (U_i, S_i) \}_{i=1}^t$, where $U_t$ is the user utterance and $S_t$ the system response. Along with the dialogue dataset, suppose we have a knowledge corpus $\{K_m \}_{m=1}^M$, where $K_m$ refers to a piece of knowledge (e.g., a sentence from Wikipedia).

Given an input $X_t = (\mathcal{D}_{t-1}, U_t)$, we aim to learn a model $f_\Theta$ to generate a knowledgeable response $\Tilde{S}_t$. Existing works frame this task as retrieving related knowledge $K_t$ for augmented input: $\Tilde{S}_t=f_\Theta(X_t, K_t)$.
Here, we propose to bypass the retrieval process by adding knowledge into the model parameters $\Theta$ to generate a response solely based on dialogue history: $\Tilde{S}_t = f_\Theta(X_t)$.

\subsection{\knowexpert~Architecture}
\label{sec:knowexpert}
\knowexpert~is composed of two components: a GPT-2 with lightweight adapters and a contextual topic model, as depicted in Figure~\ref{fig:architecture}. Inspired by~\citet{peinelt2020tbert}, the topic model is introduced to evoke knowledge stored in the GPT-2 guided by the topic information during response generation.

\paragraph{GPT-2 with Adapters}
To incorporate knowledge, we insert lightweight adapters~\cite{bapna2019simple} into each GPT-2 layer. The adapter has a two-linear-layer structure, which enables fast adaptation to targets.
Given the hidden representation of the GPT-2 layer $i$, denoted as $H_i \in \mathbb{R}^{j \times h}$, where $h$ and $j$ are the hidden dimension and the current generation step, respectively, the adapter can be formulated as
\begin{equation*}
    \mathtt{A}_\theta(H_i)=\mathtt{ReLU}(\mathtt{LN}(H_i){W_i}^{hd}){W_i}^{dh} + H_i,
\end{equation*}
where ${W_i}^{hd} \in \mathbb{R}^{h \times d}$ and ${W_i}^{dh} \in \mathbb{R}^{d \times h}$ stand for the trainable parameters in $\theta$, $\mathtt{LN}(\cdot)$ is layer normalization \citep{ba2016layer}, and $d$ is the bottleneck dimension. 
Here, we insert $L$ knowledge adapters parameterized as $ \{ \theta_{E_l} \}_{l=1}^L$ where each serves as a knowledge expert in a certain topic domain. 

\paragraph{Topic Modeling}
In~\knowexpert, a topic model is used to inform GPT-2 with more relevant ``topics'' during response generation so as to induce more context-appropriate knowledge. The topic model is trained to cluster the training knowledge corpus into a pre-defined number ($L$) of topic clusters. While any sort of topic model can be used, we adopt a contextual topic model (CTM) which outperforms traditional topic models~\citep{bianchi2021pretraining}. The CTM combines pre-trained Sentence-Transformers embedding representations~\citep{reimers2019sentence} with a neural topic model, Neural-ProdLDA~\citep{srivastava2017autoencoding}, which takes advantage of Bag of Words (BoW) for more coherent representation.

Once trained, given an input sequence, the topic model outputs a $L$-dimension vector, which is its probability distribution of the pre-clustered topics. By taking the dialogue history as inputs, these probabilities are utilized as the similarity weights $\mathbf{w} = (w_1, w_2, ..., w_L)$ over the knowledge experts to compute the weighted sum of their hidden states, as shown in Figure~\ref{fig:architecture}. We utilize $\mathbf{w}$ under two different settings: (i) the \textbf{W}eighted-sum setting where we weighted-sum the outputs from each knowledge expert when passing the hidden state to the next GPT-2 layer, and (ii) the \textbf{O}ne-hot setting where we only consider the output of the knowledge expert with the largest weight. The models trained under these two settings are denoted as \textbf{KnowExpert$_\mathrm{w}$} and \textbf{KnowExpert$_\mathrm{o}$}, respectively. 

\subsection{Learning Procedure}
\label{sec:training_procedure}
Our training follows a three-step paradigm (Figure~\ref{fig:pipeline}). In each step, each component of \knowexpert~is trained separately, which mimics human behavior during conversations referring to knowledge learned previously\cite{Tuckute2021.05.28.446230}. 

\textbf{(i) Topic Modeling Training.} We use knowledge sentences of the knowledge corpus in plain text format to train the CTM, with the pre-trained Sentence-Transformers frozen. For better guidance during training, we predict the topic distribution $\mathbf{w}$ using a concatenation of the dialogue history and the response. (We also tried other input combinations, but we achieve the best performance with the current one.)
During inference, however, this scheme cannot be applied due to the absence of responses. Thus, we further fine-tune the Sentence-Transformer inside the CTM to deal with the absence of responses. In other words, we fine-tune the Sentence-Transformer model to produce the sentence embedding of the given dialogue history as similar to the sentence embedding of the concatenation mentioned above. We leverage the mean squared error (MSE) loss to evaluate the difference between two sentence embeddings and provide the model with supervison signals.

\textbf{(ii) Knowledge Expert Training.} We train a set of $L$ topic-specific knowledge adapters inserted into the frozen backbone GPT-2 with the knowledge corpus to generate a knowledge sentence. The adapters are independently trained to minimize the negative log-likelihood over the knowledge corpus of the corresponding topic:
\begin{equation*}
    \mathcal{L}_{\mathcal{K}^l} = - \sum_{k \in \mathcal{K}^l}\sum_{1 \leq i \leq |k|} \log p(k_i|k_{<i}),
\end{equation*}
where $k_i$ is the $i$th token of a knowledge sentence in topic $\mathcal{K}^l$.

Differently to general pre-training, we expect to leverage the pretraining process on the knowledge experts to benefit the KGD task. Under this case, dialogue-oriented training is required~\citep{xu2021dialogue}. Motivated by this, we convert the format of knowledge sentences from plain text to a pseudo-conversational style to reduce the gap between knowledge expert training and task adaptation. The procedure of conversion is depicted in Figure~\ref{fig:pseudo}. 

First, we split a document of the knowledge corpus (e.g., a Wikipedia article) into sentences, and make each sentence a single utterance. Then, we randomly select 20\% of utterances and replace them with the nearest selected utterance in each dialogue to avoid the adapters over-fitting to a specific order of the knowledge sentences. The replacement is done dynamically for every epochs. Adding the token type embeddings and special tokens between knowledge sentences, we treat the knowledge sentences for knowledge expert training in the same way as the dialogues for task adaptation. Note that we make each knowledge sentence act as a system utterance and a user utterance respectively so as to ensure that each is trained as a system utterance.

\textbf{(iii) Task Adaptation.}
In the task adaptation step using the dialogue dataset, the whole GPT-2 model, except the inserted knowledge experts, is fine-tuned to generate a knowledgeable response:
\begin{equation*}
    \mathcal{L}_{\mathrm{Task}} = - \sum_{1 \leq n \leq N}\sum_{1 \leq i \leq j} \log p(s_i^{n}|s_{<i}^{n},X_t^n),
\end{equation*}
where each response is denoted as $\Tilde{S}_t^n = \{ s_i^n \}_{i=0}^j$. In this process, the number of trainable parameters is the same as that of the original GPT-2 model.

\begin{table*}[t]
\centering
\resizebox{\linewidth}{!}{
\begin{tabular}{llcccccccccc}
\toprule
\multicolumn{2}{l}{\multirow{2}{*}{\textbf{Model}}} & \multicolumn{4}{c}{WoW Seen}  & \multicolumn{4}{c}{WoW Unseen}  & \multicolumn{2}{c}{CMU\_DoG}    \\ 
\cmidrule(lr){3-6} \cmidrule(lr){7-10} \cmidrule(lr){11-12}
\multicolumn{2}{c}{}  & \multicolumn{1}{c}{PPL$\downarrow$} & \multicolumn{1}{c}{F1$\uparrow$} & \multicolumn{1}{c}{Dist-1$\uparrow$} & \multicolumn{1}{c}{Dist-2$\uparrow$} & \multicolumn{1}{c}{PPL$\downarrow$} & \multicolumn{1}{c}{F1$\uparrow$} & \multicolumn{1}{c}{Dist-1$\uparrow$} & \multicolumn{1}{c}{Dist-2$\uparrow$} & \multicolumn{1}{c}{PPL$\downarrow$} & \multicolumn{1}{c}{F1$\uparrow$} \\ \midrule
\multirow{5}{*}{\begin{tabular}[c]{@{}l@{}}Retrieval-based\\ Approach\end{tabular}} 
 & DRD & 23.0 & \underline{18.0} & - & - & 25.6  & \underline{16.5} & - & - & 54.4 & \underline{10.7} \\
 & ZRGKG & 40.4 & \underline{18.7} & \underline{5.4} & \underline{22.5} & 41.5 & 18.6 & \underline{3.4} & \underline{15.6} & 53.5 & \underline{12.5} \\
 & GPT-2$_\mathrm{{trunc}}$ & 14.6 & \underline{18.7} & - & - & 16.9 & 18.3 & - & - & 18.6 & \underline{10.8} \\
 & KnowledGPT & 19.2 & \textbf{22.0} & 8.9 & 36.2 & 22.3 & \textbf{20.5} & 6.0 & 23.8 & 20.6 & \textbf{13.5} \\\midrule
\multirow{5}{*}{\begin{tabular}[c]{@{}l@{}}Retrieval-free\\ Approach\end{tabular}} & GPT-2$_\mathrm{f}$ & 18.8 & 17.0 & 4.9 & 21.1 & 21.0 & 16.3 & 3.9 & 16.8 & 17.8 & 11.8 \\
& KE-Blender$^{\dagger}$  & 15.5 & 17.0 & - & - & 18.4 & \textbf{16.7} & - & - & - & - \\
& $\text{KnowExpert}_\mathrm{w}$+causal & 15.2 & 18.4 & 6.4 & 26.4 & 20.0 & 16.6 & 4.9 & 20.4 & 16.8 & 12.1 \\
& $\text{KnowExpert}_\mathrm{o}$ (ours) & 16.0 & 18.4 & 6.6 & 27.2 & 21.2 & 16.6 & \textbf{5.2} & \textbf{21.6} & 17.8 & 12.1 \\
& $\text{KnowExpert}_\mathrm{w}$ (ours) & 15.3 & \textbf{18.7} & \textbf{6.8} & \textbf{27.9} & 20.1 & \textbf{16.7} & \textbf{5.2} & 21.2 & 17.2 & \textbf{12.5} \\\bottomrule
\end{tabular}
}
\caption{Automatic evaluation results ($L=4$). PPL is short for Perplexity; F1 refers to the unigram-F1 score between the generated and gold responses; Dist-1/2 denotes uni-gram and bi-gram distinct metrics. We highlight the best results for each group in \textbf{bold}. We also \underline{underline} the cases when our proposed \knowexpert~outperforms the retrieval-based models. $^{\dagger}$Although KE-Blender is not a retrieval-free model, we present its reported inference performance without the knowledge inputs.}
\label{tab:results}
\end{table*}

\section{Experiments}
\subsection{Datasets}
\label{subsec:data}
We conduct experiments on two datasets: Wizard of Wikipedia (WoW)~\cite{dinan2018wizard} and CMU Document Grounded Conversations (CMU\_DoG)~\cite{zhou2018dataset}. In the training process, we collect all the knowledge sentences provided by the WoW and CMU\_DoG datasets to build a knowledge corpus with 117,495 articles. 

\begin{table}[t]
\centering
\resizebox{.49\textwidth}{!}{
\begin{tabular}{lcccc}
\toprule
\multicolumn{1}{l}{Winning Rate (\%)} & \multicolumn{2}{c}{WoW Seen} & \multicolumn{2}{c}{WoW Unseen} \\
\cmidrule(lr){1-1} \cmidrule(lr){2-3} \cmidrule(lr){4-5}
\multicolumn{1}{l}{Models}               & \multicolumn{1}{c}{Info.} & \multicolumn{1}{c}{Human.} & \multicolumn{1}{c}{Info.} & \multicolumn{1}{c}{Human.} \\ \midrule
KnowExpert$_\mathrm{w}$ vs. GPT-2$_\mathrm{f}$ & \textbf{57.68} & 48.69 & \textbf{59.26} & \textbf{56.13} \\ 
KnowExpert$_\mathrm{o}$ vs. GPT-2$_\mathrm{f}$ & \textbf{64.46} & \textbf{54.42} & \textbf{55.88} & \textbf{53.67} \\
\bottomrule
\end{tabular}
}
\caption{Human evaluation results in terms of the winning rate of our model over the GPT-2$_\mathrm{f}$ baseline for \textit{Informativeness} and \textit{Humanness}. A significance pair-wise t-test is conducted and the results in bold are significantly better than those from the baseline model ($p < 0.05$). } 
\label{tab:human_eval}
\end{table}

\subsection{Training Details}
\paragraph{Topic Modeling.}
For preprocessing, we limit the vocaburary size for BoW to 20000. The number of topic clusters $L$ is set as 4.  We use the frozen RoBERTa (125M) model pre-trained with the NLI datasets~\citep{conneau2017supervised} and STS Benchmark~\citep{cer2017semeval} provided by~\citet{wolf-etal-2020-transformers} as the Sentence-Transformer inside the CTM. The CTM is trained with the Adam optimizer~\cite{kingma2015adam} with $\beta_1 = 0.9, \beta_2 = 0.999$, and a learning rate of $2\mathrm{e}{-3}$. For further fine-tuning of RoBERTA, we apply the Adam optimizer with the same $\beta_1, \beta_2$ and a learning rate of $1\mathrm{e}{-6}$ with a linear scheduler.

\paragraph{Knowledge Expert Training.}
We utilize the CTM model to split the knowledge corpus mentioned above into $\mathcal{L}$ clusters for training the corresponding $\mathcal{L}$ knowledge experts. In the experiments, we utilize the pre-trained GPT-2 (117M) model provided by~\citet{wolf-etal-2020-transformers}. The adapter bottleneck dimension $d$ is set to be 768 for the knowledge adapters. All the adapters are learned with the Adam optimizer with $\beta_1 = 0.9, \beta_2 = 0.999$. The learning rate is set to be $1\mathrm{e}{-4}$ for knowledge expert training with a linear scheduler, and the knowledge experts are trained with 50 epochs.

\paragraph{Task Adaptation.} 
For task adaptation, we keep the same hyper-parameter setting as in knowledge expert training, while the learning rate is set as $1\mathrm{e}{-5}$. The maximum number of training epochs is set as 50 with a linear learning rate scheduler and the patience for early stopping as 5. We employ a greedy search in decoding responses. Also noted that, each experiment mentioned above is conducted on a single RTX 2080 Ti GPU. 


\begin{table}[t]
\centering
\resizebox{.45\textwidth}{!}{
\begin{tabular}{cccccc}
\toprule
\multicolumn{1}{l}{\multirow{2}{*}{\begin{tabular}[c]{@{}l@{}}\# of\\
Clus.\end{tabular}}} & \multicolumn{2}{c}{WoW Seen}                     & \multicolumn{2}{c}{WoW Unseen} & \multicolumn{1}{c}{Average}                   \\
\cmidrule(lr){2-3} \cmidrule(lr){4-5} \cmidrule(lr){6-6} 
\multicolumn{1}{l}{} & \multicolumn{1}{c}{PPL$\downarrow$} & \multicolumn{1}{c}{F1$\uparrow$} & \multicolumn{1}{c}{PPL$\downarrow$} & \multicolumn{1}{c}{F1$\uparrow$} & \multicolumn{1}{c}{F1$\uparrow$} \\ \midrule
  4 & \textbf{15.95} & \textbf{18.41} & 21.18 & 16.61 & \textbf{17.51} \\
  8 & 16.22 & 18.14 & 21.21 & 16.58 & 17.36 \\
 16 & 16.43 & 18.05 & \textbf{21.12} & \textbf{16.76} & 17.41 \\ \bottomrule
\end{tabular}
}
\caption{Effects of the number of topic clusters. We present the results when setting the number of pre-defined topic clusters as 4, 8 and 16 while utilizing one-hot knowledge adapters (\knowexpert$_\mathrm{o}$) to keep the same number of parameters in the models.}
\label{tab:cluster_num}
\end{table}

\begin{table*}[t!]
\centering
\resizebox{\textwidth}{!}{
\begin{tabular}{@{}ll|ll@{}}
\toprule
\multicolumn{2}{l|}{\multirow{3}{*}{Context}} &
  \multicolumn{2}{l}{User: Orc.} \\
\multicolumn{2}{c|}{} &
  \multicolumn{2}{l}{System: Orcs are cool fictional humanoid beings.} \\
\multicolumn{2}{c|}{} &
  \multicolumn{2}{l}{User: Yeah, I’ve seen them in a lot of things like magic and dnd.} \\ \midrule
\multicolumn{1}{l|}{\multirow{8}{*}{\begin{tabular}[c]{@{}l@{}}Generated \\ responses \\ with single \\ knowledge \\ expert in \\ KnowExpert$_\mathrm{w}$ \\ ($L=4$)\end{tabular}}} &
  \multirow{2}{*}{Expert 1} &
  \multicolumn{1}{l|}{Do you know about the orcs? They are native to the \underline{Italian peninsula}.} &
  \multirow{2}{*}{\ding{55}} \\
\multicolumn{1}{l|}{} &
   &
  \multicolumn{1}{l|}{\textit{\textbf{Topics of Cluster 1}: east, river, south, state, city, area, island, ...}} &
   \\ \cmidrule(l){2-4} 
\multicolumn{1}{l|}{} &
  \multicolumn{1}{c|}{\multirow{2}{*}{Expert 2}} &
  \multicolumn{1}{l|}{They are a subgenre of “art games” that are a subgenre of video games.
  \quad} &
  \multirow{2}{*}{\ding{55}} \\
\multicolumn{1}{l|}{} &
  \multicolumn{1}{c|}{} &
  \multicolumn{1}{l|}{\textit{\textbf{Topics of Cluster 2}: rock, band, music, album, football, single, ...}} &
   \\ \cmidrule(l){2-4} 
\multicolumn{1}{l|}{} &
  \multirow{2}{*}{Expert 3} &
  \multicolumn{1}{l|}{Orcs are cool, they are a subspecies of \underline{elves in the warcraft universe}.} &
  \multirow{2}{*}{\ding{51}} \\
\multicolumn{1}{l|}{} &
   &
  \multicolumn{1}{l|}{\textit{\textbf{Topics of Cluster 3}: fiction, story, characters, novel, film, stars, ...}} &
   \\ \cmidrule(l){2-4} 
\multicolumn{1}{l|}{} &
  \multirow{2}{*}{Expert 4} &
  \multicolumn{1}{l|}{They are a legendary race that are native to the americas.} &
  \multirow{2}{*}{\ding{55}} \\
\multicolumn{1}{l|}{} &
   &
  \multicolumn{1}{l|}{\textit{\textbf{Topics of Cluster 4}: bon, bucks, rutgers, canberra, ivy, nets, ...}} &
   \\ \midrule
\multicolumn{2}{l|}{KnowExpert$_\mathrm{w}$} &
  \multicolumn{2}{l}{\begin{tabular}[c]{@{}l@{}}They are a fictional humanoid creature from the "dungeons \& dragons" fantasy \\ roleplaying game.\end{tabular}} \\ \bottomrule
\end{tabular}
}
\caption{Case study on the effect of different knowledge experts in \texttt{KnowExpert$_\mathrm{w}$} ($L=4$). \textit{Expert 1/2/3/4} denotes the generated responses with the same context with \texttt{KnowExpert$_\mathrm{w}$} using different knowledge experts separately on the WoW test seen set. Along with the generated responses, we also show the topic keywords of each cluster extracted with the topic model in \S~\ref{sec:knowexpert}. In this example, Expert 3 is more related to the topic of the dialogue context.}
\label{tab:case_study}
\end{table*}

\subsection{Baselines}
We selected baseline models which follow the retrieval-encode schema, based on the relevance to our experimental settings: (i) \textbf{DRD}~\citep{zhao2019low} intends to combat low-resource settings with pre-training techniques; (ii) \textbf{ZRGKG}~\citep{li2020zero-resource} explores the response generation problem without leveraging the matching annotations between the context-response and the knowledge sentences during training; (iii) \textbf{GPT-2$_\mathrm{trunc}$}~\citep{zhao2020knowledge} randomly ranks the provided knowledge sentences and directly concatenates them with the dialogue context as inputs, while truncating the part exceeding the maximum input length; (iv) \textbf{KnowledGPT}~\citep{zhao2020knowledge} exploits pre-trained LMs as both a knowledge selection module and a response generation module which are optimized jointly; (v) \textbf{KE-Blender}~\citep{cui2021knowledge} proposes knowledge-enhanced finetuning during training to better handle the unseen entities in the dialogue history. KE-Blender is not a retrieval-free model, but we focus on the case of no knowledge inputs during inference for KE-Blender, which is similar to our settings.

As an additional baseline for comparison among the solely generation-based approaches, we fine-tune the whole GPT-2 model to generate responses given dialogue contexts, without accessing an explicit knowledge corpus (\textbf{GPT-2$_\mathrm{f}$}). To evaluate the effect of dialogue-oriented training for knowledge experts, we train the knowledge adapters with GPT-2-style causal pre-training and keep the other settings unchanged. The corresponding model is denoted as \textbf{KnowExpert$_\mathrm{w}$+causal}.

\subsection{Evaluation and Model Selection}
\paragraph{Automatic Metrics}
Following~\citet{dinan2018wizard}, we present the perplexity (PPL) of generating the gold responses and uni-gram F1 as automatic evaluation metrics. The uni-gram F1 metric is implemented with the ParlAI~\footnote{\url{https://github.com/facebookresearch/ParlAI}} package. 
In addition, we also evaluate the uni-gram and bi-gram diversity of the generated response with the corpus-level DISTINCT~\citep{li2016diversity} metric.

\paragraph{Human Evaluation}
In addition to the automatic evaluation, we conduct human evaluation over the generated responses from two aspects: \textit{Informativeness (Info.)} and \textit{Humanness (Human.)}. ``Info.'' evaluates how knowledgeable the generated responses are, based on the amount of new information introduced into the conversations and the factuality of the responses, while ``Human.'' is used for evaluating the fluency and the coherence of the generated responses. 

A/B testing is utilized to compare our proposed framework, KnowExpert$_\mathrm{w}$ and KnowExpert$_\mathrm{o}$, with the GPT-2$_\mathrm{f}$ baseline on the WoW dataset. For each comparison, the same context and two options generated by the models in comparison are shown to the annotators. Each comparison requires three judgments, and 100 data samples are randomly selected from each domain. We conduct a human evaluation using a crowd-sourcing platform offered by Amazon Mechanical Turk.\footnote{\url{https://www.mturk.com}} We ensure that each sample is evaluated by three annotators. Further details and annotator instructions are included in Appendix~\ref{sec:human_eval_appd}.

\paragraph{Model Selection} In the training procedure, we have different criteria for selecting models for the three training steps: In (i) and (ii), we train the corresponding model for a specific number of epochs; in (iii), the model is selected according to the sum of the PPLs on the seen and unseen validation sets.

\subsection{Results}

Table~\ref{tab:results} reports the automatic evaluation results. 
The improvements over the baseline model GPT-2$_\mathrm{f}$ demonstrate the effectiveness of our proposed framework. In this task, KnowExpert$_\mathrm{w}$ performs comparably with the retrieval-based baselines, especially on the seen domain, without using either retrieval or any explicit extra knowledge input in the inference process. Compared with the KE-Blender model under the retrieval-free setting, \knowexpert shows a significant advantage on the WoW seen. In addition, KnowExpert$_\mathrm{w}$ also shows consistently better performance over KnowExpert$_\mathrm{o}$. Without dialogue-oriented training, the performance of the proposed model ($\text{KnowExpert}_\mathrm{w}$+causal) drops even below tha of the model with the one-hot setting, which shows the importance of dialogue-oriented training.
Despite the improvements over the baseline model, we also observe a performance gap between the seen and unseen domains, which requires future work.

\begin{figure*}[t!]
    \centering
    \resizebox{\textwidth}{!}{  
        \includegraphics{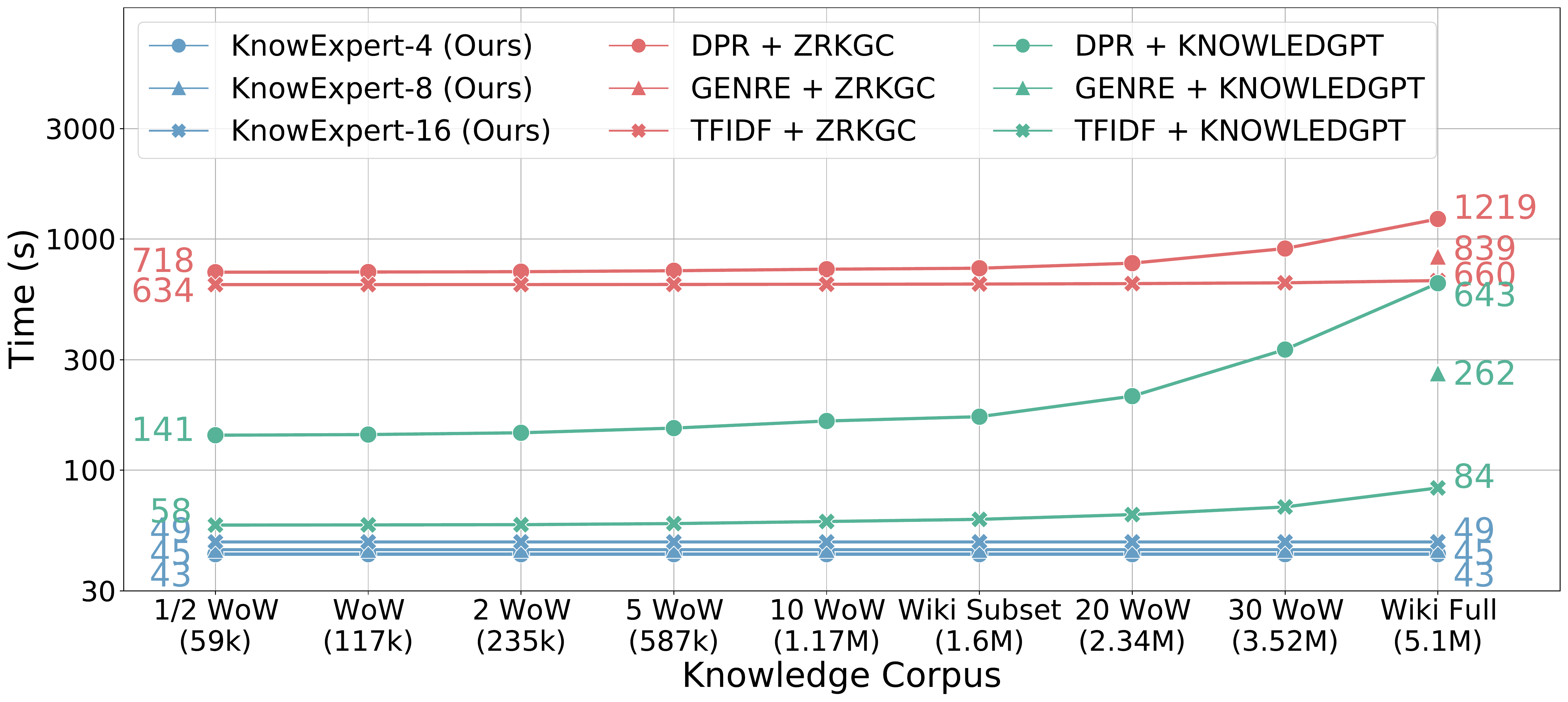}
    }
    \caption{Inference efficiency of our approach generating 100 samples. We show the time on a logarithmic scale and the knowledge corpus sizes in ascending order. In the figure above, we demonstrate the overall end-to-end inference time of our method with the generation length of 23 (average response length in WoW dataset).}
    \label{fig:efficiency}
\end{figure*}


Table~\ref{tab:human_eval} shows the human evaluation results in terms of the winning rate for Info. and Human.
The results indicate that the introduction of the knowledge experts brings the GPT-2 model a significant improvement in generating a more informative response, without hurting the fluency and coherence of the generation under the weighted-sum setting. However, when using the knowledge experts under a one-hot setting, the improvement is not as large as that of the weighted-sum one on the unseen domain, which follows the results of the automatic metrics.   

\subsection{Effects of Number of Topic Clusters}
\label{sec:num_cluster}
The number of topic clusters is an important hyperparameter since it crucially impacts the quality of topic modeling and knowledge expert training. Because of the nature of the WoW and CMU\_DoG datasets, we conduct experiments with $L=4, 8, 16$. In Table~\ref{tab:topic_keywords} in the Appendix, we show in detail the most frequent words for each topic cluster with different numbers of topic clusters. For example, Cluster 2 when $L=8$ is strongly related to the movie domain.
As shown in Table~\ref{tab:cluster_num}, we select $L=4$ since it achieves the best average F1 on two WoW test sets. 



\subsection{Case Study}
\label{sec:case_study}
We leverage different knowledge experts in a one-hot manner, generating responses with only one knowledge expert and the same dialogue history to study what each knowledge expert captures.
As shown in Table~\ref{tab:case_study}, the responses generated with different knowledge experts tend to lean into different cluster topics with the same context. We also provide another example in Table~\ref{tab:case_study2}. Some selected keywords are shown below, and more topic keywords are listed in Table~\ref{tab:topic_keywords} in the Appendix. Comparing the responses with the listed topic keywords, our knowledge experts tend to focus on the topics to which the knowledge documents they are trained on belong. For example, with the same context, Expert 2 is leaning into the music domain as Cluster 2 is strongly related to music, while Expert 3 relates more to the fiction topics, which align with the topic in Cluster 3. In addition, the shown cases also support the observation from Table~\ref{tab:results} that the mixture-of-experts approach ensures a better model performance. The generated response of KnowExpert$_\mathrm{w}$ is more on-topic and accurate thanks to leveraging the weighted sum of the experts. The above findings indicate that the proper ensemble of experts also helps the response generation.

Although the generated responses appear to be knowledgeable and fluent, they frequently raise an issue of factual correctness; for example, ``Orcs'' are not directly related to the ``Italian peninsula''. We also observe that a knowledge expert whose topics are more similar to the topic of the dialogue tends to generate more factual responses.

\subsection{Inference Efficiency}
%
We evaluate the response generation inference time of \knowexpert~and two other retrieval-based baselines: ZRGKG and KnowledGPT. In addition to the time to generate responses, we also consider the time required for retrieving knowledge candidates from knowledge corpora of different sizes against the time required for topic modeling in \knowexpert. We take the retrieval methods TF-IDF, DPR~\citep{karpukhin2020dense}, and GENRE~\citep{decao2020autoregressive} for comparison. To have a fair comparison with our approach, we measure the end-to-end inference time by summing the time for retrieval and response generation. The generation length is pre-defined as the average response length in the WoW dataset. We randomly sample 100 instances from WoW seen and unseen test set and average the inference time of 10 trials. The detailed configuration is listed in Table C1 in the Appendix.

As shown in Figure~\ref{fig:efficiency}, \knowexpert~requires the least computing time and keeps a constant computational cost, regardless of the size of the knowledge corpus. This is because our topic modeling requires a constant computational cost, while that of TF-IDF or DPR incurs an increasing cost as the size of the knowledge corpus increases. Additionally, our model does not require a large external corpus during the inference time. These results suggest that our model is suitable for deployment in resource-limited platforms, such as in the on-device setting. 

\section{Conclusion}
We propose \knowexpert, a retrieval-free framework for the KGD task. \knowexpert is the first attempt to tackle the challenge of injecting knowledge into the model parameters and leveraging it for the KGD task. We leverage light-weight adapters as knowledge experts, then train the backbone model to take advantage of them for response generation. By these means, our method can generate more knowledgeable responses without an explicit retrieval step compared to our baseline model. By bypassing the retrieval step over the knowledge corpus, the inference efficiency of the model is improved. Experimental results show that \knowexpert~performs comparably with some retrieval-based models, demonstrating the promise of our proposed research direction. 

\section*{Acknowledgement}

We would thank Zhaojiang Lin, Peng Xu for discussing the project and the anonymous reviewers for the insightful reviews and constructive feedback.
This work has been partially supported by the HKJCCT21EG01 of the Hong Kong Jockey Club, and School of Engineering Ph.D. Fellowship Award, the Hong Kong University of Science and Technology.

\bibliography{anthology,custom}

\begin{thebibliography}{44}
\expandafter\ifx\csname natexlab\endcsname\relax\def\natexlab#1{#1}\fi

\bibitem[{Ba et~al.(2016)Ba, Kiros, and Hinton}]{ba2016layer}
Lei~Jimmy Ba, Jamie~Ryan Kiros, and Geoffrey~E. Hinton. 2016.
\newblock \href {http://arxiv.org/abs/1607.06450} {Layer normalization}.
\newblock \emph{CoRR}, abs/1607.06450.

\bibitem[{Bapna and Firat(2019)}]{bapna2019simple}
Ankur Bapna and Orhan Firat. 2019.
\newblock Simple, scalable adaptation for neural machine translation.
\newblock In \emph{Proceedings of the 2019 Conference on Empirical Methods in
  Natural Language Processing and the 9th International Joint Conference on
  Natural Language Processing (EMNLP-IJCNLP)}, pages 1538--1548.

\bibitem[{Bianchi et~al.(2021)Bianchi, Terragni, and
  Hovy}]{bianchi2021pretraining}
Federico Bianchi, Silvia Terragni, and Dirk Hovy. 2021.
\newblock Pre-training is a hot topic: Contextualized document embeddings
  improve topic coherence.
\newblock In \emph{ACL}.

\bibitem[{Brown et~al.(2020)Brown, Mann, Ryder, Subbiah, Kaplan, Dhariwal,
  Neelakantan, Shyam, Sastry, Askell et~al.}]{brown2020language}
Tom~B Brown, Benjamin Mann, Nick Ryder, Melanie Subbiah, Jared Kaplan, Prafulla
  Dhariwal, Arvind Neelakantan, Pranav Shyam, Girish Sastry, Amanda Askell,
  et~al. 2020.
\newblock Language models are few-shot learners.
\newblock \emph{arXiv preprint arXiv:2005.14165}.

\bibitem[{Cao et~al.(2021)Cao, Izacard, Riedel, and
  Petroni}]{decao2020autoregressive}
Nicola~De Cao, Gautier Izacard, Sebastian Riedel, and Fabio Petroni. 2021.
\newblock \href {https://openreview.net/forum?id=5k8F6UU39V} {Autoregressive
  entity retrieval}.
\newblock In \emph{International Conference on Learning Representations}.

\bibitem[{Cao et~al.(2020)Cao, Trivedi, Balasubramanian, and
  Balasubramanian}]{cao2020deformer}
Qingqing Cao, Harsh Trivedi, Aruna Balasubramanian, and Niranjan
  Balasubramanian. 2020.
\newblock \href {https://doi.org/10.18653/v1/2020.acl-main.411} {{D}e{F}ormer:
  Decomposing pre-trained transformers for faster question answering}.
\newblock In \emph{Proceedings of the 58th Annual Meeting of the Association
  for Computational Linguistics}, pages 4487--4497, Online. Association for
  Computational Linguistics.

\bibitem[{Cer et~al.(2017)Cer, Diab, Agirre, Lopez-Gazpio, and
  Specia}]{cer2017semeval}
Daniel Cer, Mona Diab, Eneko Agirre, I{\~n}igo Lopez-Gazpio, and Lucia Specia.
  2017.
\newblock \href {https://doi.org/10.18653/v1/S17-2001} {{S}em{E}val-2017 task
  1: Semantic textual similarity multilingual and crosslingual focused
  evaluation}.
\newblock In \emph{Proceedings of the 11th International Workshop on Semantic
  Evaluation ({S}em{E}val-2017)}, pages 1--14, Vancouver, Canada. Association
  for Computational Linguistics.

\bibitem[{Chen et~al.(2020)Chen, Meng, Li, Chen, Xu, Xu, and
  Zhou}]{chen2020bridging}
Xiuyi Chen, Fandong Meng, Peng Li, Feilong Chen, Shuang Xu, Bo~Xu, and Jie
  Zhou. 2020.
\newblock Bridging the gap between prior and posterior knowledge selection for
  knowledge-grounded dialogue generation.
\newblock In \emph{Proceedings of the 2020 Conference on Empirical Methods in
  Natural Language Processing (EMNLP)}, pages 3426--3437.

\bibitem[{Conneau et~al.(2017)Conneau, Kiela, Schwenk, Barrault, and
  Bordes}]{conneau2017supervised}
Alexis Conneau, Douwe Kiela, Holger Schwenk, Lo{\"\i}c Barrault, and Antoine
  Bordes. 2017.
\newblock \href {https://doi.org/10.18653/v1/D17-1070} {Supervised learning of
  universal sentence representations from natural language inference data}.
\newblock In \emph{Proceedings of the 2017 Conference on Empirical Methods in
  Natural Language Processing}, pages 670--680, Copenhagen, Denmark.
  Association for Computational Linguistics.

\bibitem[{Cui et~al.(2021)Cui, Wu, Liu, and Zhang}]{cui2021knowledge}
Leyang Cui, Yu~Wu, Shujie Liu, and Yue Zhang. 2021.
\newblock Knowledge enhanced fine-tuning for better handling unseen entities in
  dialogue generation.
\newblock In \emph{Proceedings of the 2021 Conference on Empirical Methods in
  Natural Language Processing}, pages 2328--2337.

\bibitem[{Dinan et~al.(2019)Dinan, Roller, Shuster, Fan, Auli, and
  Weston}]{dinan2018wizard}
Emily Dinan, Stephen Roller, Kurt Shuster, Angela Fan, Michael Auli, and Jason
  Weston. 2019.
\newblock \href {https://openreview.net/forum?id=r1l73iRqKm} {Wizard of
  wikipedia: Knowledge-powered conversational agents}.
\newblock In \emph{International Conference on Learning Representations}.

\bibitem[{Jiao et~al.(2020)Jiao, Yin, Shang, Jiang, Chen, Li, Wang, and
  Liu}]{jiao2020tinybert}
Xiaoqi Jiao, Yichun Yin, Lifeng Shang, Xin Jiang, Xiao Chen, Linlin Li, Fang
  Wang, and Qun Liu. 2020.
\newblock \href {https://doi.org/10.18653/v1/2020.findings-emnlp.372}
  {{T}iny{BERT}: Distilling {BERT} for natural language understanding}.
\newblock In \emph{Findings of the Association for Computational Linguistics:
  EMNLP 2020}, pages 4163--4174, Online. Association for Computational
  Linguistics.

\bibitem[{Karpukhin et~al.(2020)Karpukhin, Oguz, Min, Lewis, Wu, Edunov, Chen,
  and Yih}]{karpukhin2020dense}
Vladimir Karpukhin, Barlas Oguz, Sewon Min, Patrick Lewis, Ledell Wu, Sergey
  Edunov, Danqi Chen, and Wen-tau Yih. 2020.
\newblock Dense passage retrieval for open-domain question answering.
\newblock In \emph{Proceedings of the 2020 Conference on Empirical Methods in
  Natural Language Processing (EMNLP)}, pages 6769--6781.

\bibitem[{Kim et~al.(2019)Kim, Ahn, and Kim}]{kim2019sequential}
Byeongchang Kim, Jaewoo Ahn, and Gunhee Kim. 2019.
\newblock Sequential latent knowledge selection for knowledge-grounded
  dialogue.
\newblock In \emph{International Conference on Learning Representations}.

\bibitem[{Kingma and Ba(2015)}]{kingma2015adam}
Diederik~P. Kingma and Jimmy Ba. 2015.
\newblock \href {http://arxiv.org/abs/1412.6980} {Adam: {A} method for
  stochastic optimization}.
\newblock In \emph{3rd International Conference on Learning Representations,
  {ICLR} 2015, San Diego, CA, USA, May 7-9, 2015, Conference Track
  Proceedings}.

\bibitem[{Lauscher et~al.(2020)Lauscher, Majewska, Ribeiro, Gurevych, Rozanov,
  and Glava{\v{s}}}]{lauscher2020common}
Anne Lauscher, Olga Majewska, Leonardo~FR Ribeiro, Iryna Gurevych, Nikolai
  Rozanov, and Goran Glava{\v{s}}. 2020.
\newblock Common sense or world knowledge? investigating adapter-based
  knowledge injection into pretrained transformers.
\newblock \emph{arXiv preprint arXiv:2005.11787}.

\bibitem[{Li et~al.(2016)Li, Galley, Brockett, Gao, and
  Dolan}]{li2016diversity}
Jiwei Li, Michel Galley, Chris Brockett, Jianfeng Gao, and William~B Dolan.
  2016.
\newblock A diversity-promoting objective function for neural conversation
  models.
\newblock In \emph{Proceedings of the 2016 Conference of the North American
  Chapter of the Association for Computational Linguistics: Human Language
  Technologies}, pages 110--119.

\bibitem[{Li et~al.(2020)Li, Xu, Wu, ZHAO, Zhao, and Tao}]{li2020zero-resource}
Linxiao Li, Can Xu, Wei Wu, YUFAN ZHAO, Xueliang Zhao, and Chongyang Tao. 2020.
\newblock \href
  {https://proceedings.neurips.cc/paper/2020/file/609c5e5089a9aa967232aba2a4d03114-Paper.pdf}
  {Zero-resource knowledge-grounded dialogue generation}.
\newblock In \emph{Advances in Neural Information Processing Systems},
  volume~33, pages 8475--8485. Curran Associates, Inc.

\bibitem[{Lian et~al.(2019)Lian, Xie, Wang, Peng, and Wu}]{lian2019learning}
Rongzhong Lian, Min Xie, Fan Wang, Jinhua Peng, and Hua Wu. 2019.
\newblock Learning to select knowledge for response generation in dialog
  systems.
\newblock In \emph{Proceedings of the 28th International Joint Conference on
  Artificial Intelligence}, pages 5081--5087. AAAI Press.

\bibitem[{Liu et~al.(2022)Liu, Patwary, Prenger, Prabhumoye, Ping, Shoeybi, and
  Catanzaro}]{liu2022multi}
Zihan Liu, Mostofa Patwary, Ryan Prenger, Shrimai Prabhumoye, Wei Ping,
  Mohammad Shoeybi, and Bryan Catanzaro. 2022.
\newblock Multi-stage prompting for knowledgeable dialogue generation.
\newblock \emph{arXiv preprint arXiv:2203.08745}.

\bibitem[{Madotto et~al.(2020)Madotto, Cahyawijaya, Winata, Xu, Liu, Lin, and
  Fung}]{madotto2020learning}
Andrea Madotto, Samuel Cahyawijaya, Genta~Indra Winata, Yan Xu, Zihan Liu,
  Zhaojiang Lin, and Pascale Fung. 2020.
\newblock Learning knowledge bases with parameters for task-oriented dialogue
  systems.
\newblock In \emph{Proceedings of the 2020 Conference on Empirical Methods in
  Natural Language Processing: Findings}, pages 2372--2394.

\bibitem[{Peinelt et~al.(2020)Peinelt, Nguyen, and Liakata}]{peinelt2020tbert}
Nicole Peinelt, Dong Nguyen, and Maria Liakata. 2020.
\newblock \href {https://doi.org/10.18653/v1/2020.acl-main.630} {t{BERT}: Topic
  models and {BERT} joining forces for semantic similarity detection}.
\newblock In \emph{Proceedings of the 58th Annual Meeting of the Association
  for Computational Linguistics}, pages 7047--7055, Online. Association for
  Computational Linguistics.

\bibitem[{Petroni et~al.(2019)Petroni, Rockt{\"a}schel, Riedel, Lewis, Bakhtin,
  Wu, and Miller}]{petroni2019language}
Fabio Petroni, Tim Rockt{\"a}schel, Sebastian Riedel, Patrick Lewis, Anton
  Bakhtin, Yuxiang Wu, and Alexander Miller. 2019.
\newblock Language models as knowledge bases?
\newblock In \emph{Proceedings of the 2019 Conference on Empirical Methods in
  Natural Language Processing and the 9th International Joint Conference on
  Natural Language Processing (EMNLP-IJCNLP)}, pages 2463--2473.

\bibitem[{Radford et~al.(2019)Radford, Wu, Child, Luan, Amodei, and
  Sutskever}]{radford2019language}
Alec Radford, Jeffrey Wu, Rewon Child, David Luan, Dario Amodei, and Ilya
  Sutskever. 2019.
\newblock Language models are unsupervised multitask learners.
\newblock \emph{OpenAI Blog, 1(8):9.}

\bibitem[{Raffel et~al.(2019)Raffel, Shazeer, Roberts, Lee, Narang, Matena,
  Zhou, Li, and Liu}]{raffel2019exploring}
Colin Raffel, Noam Shazeer, Adam Roberts, Katherine Lee, Sharan Narang, Michael
  Matena, Yanqi Zhou, Wei Li, and Peter~J Liu. 2019.
\newblock Exploring the limits of transfer learning with a unified text-to-text
  transformer.
\newblock \emph{arXiv preprint arXiv:1910.10683}.

\bibitem[{Reimers and Gurevych(2019)}]{reimers2019sentence}
Nils Reimers and Iryna Gurevych. 2019.
\newblock \href {https://doi.org/10.18653/v1/D19-1410} {Sentence-{BERT}:
  Sentence embeddings using {S}iamese {BERT}-networks}.
\newblock In \emph{Proceedings of the 2019 Conference on Empirical Methods in
  Natural Language Processing and the 9th International Joint Conference on
  Natural Language Processing (EMNLP-IJCNLP)}, pages 3982--3992, Hong Kong,
  China. Association for Computational Linguistics.

\bibitem[{Roberts et~al.(2020)Roberts, Raffel, and Shazeer}]{roberts2020much}
Adam Roberts, Colin Raffel, and Noam Shazeer. 2020.
\newblock How much knowledge can you pack into the parameters of a language
  model?
\newblock In \emph{Proceedings of the 2020 Conference on Empirical Methods in
  Natural Language Processing (EMNLP)}, pages 5418--5426.

\bibitem[{Roller et~al.(2020)Roller, Dinan, Goyal, Ju, Williamson, Liu, Xu,
  Ott, Shuster, Smith et~al.}]{roller2020recipes}
Stephen Roller, Emily Dinan, Naman Goyal, Da~Ju, Mary Williamson, Yinhan Liu,
  Jing Xu, Myle Ott, Kurt Shuster, Eric~M Smith, et~al. 2020.
\newblock Recipes for building an open-domain chatbot.
\newblock \emph{arXiv preprint arXiv:2004.13637}.

\bibitem[{Sanh et~al.(2019)Sanh, Debut, Chaumond, and
  Wolf}]{sanh2019distilbert}
Victor Sanh, Lysandre Debut, Julien Chaumond, and Thomas Wolf. 2019.
\newblock \href {http://arxiv.org/abs/1910.01108} {Distilbert, a distilled
  version of {BERT:} smaller, faster, cheaper and lighter}.
\newblock \emph{CoRR}, abs/1910.01108.

\bibitem[{Shin et~al.(2020)Shin, Razeghi, Logan~IV, Wallace, and
  Singh}]{shin2020autoprompt}
Taylor Shin, Yasaman Razeghi, Robert~L Logan~IV, Eric Wallace, and Sameer
  Singh. 2020.
\newblock Autoprompt: Eliciting knowledge from language models with
  automatically generated prompts.
\newblock \emph{arXiv preprint arXiv:2010.15980}.

\bibitem[{Shwartz et~al.(2020)Shwartz, West, Bras, Bhagavatula, and
  Choi}]{shwartz2020unsupervised}
Vered Shwartz, Peter West, Ronan~Le Bras, Chandra Bhagavatula, and Yejin Choi.
  2020.
\newblock Unsupervised commonsense question answering with self-talk.
\newblock \emph{arXiv preprint arXiv:2004.05483}.

\bibitem[{Srivastava and Sutton(2017)}]{srivastava2017autoencoding}
Akash Srivastava and Charles Sutton. 2017.
\newblock \href {https://arxiv.org/abs/1703.01488} {Autoencoding variational
  inference for topic models}.
\newblock In \emph{ICLR}.

\bibitem[{Sun et~al.(2019)Sun, Li, Wang, He, Lin, and Deng}]{sun2019fast}
Zhiqing Sun, Zhuohan Li, Haoqing Wang, Di~He, Zi~Lin, and Zhihong Deng. 2019.
\newblock \href
  {https://proceedings.neurips.cc/paper/2019/file/74563ba21a90da13dacf2a73e3ddefa7-Paper.pdf}
  {Fast structured decoding for sequence models}.
\newblock In \emph{Advances in Neural Information Processing Systems},
  volume~32. Curran Associates, Inc.

\bibitem[{Sun et~al.(2020)Sun, Yu, Song, Liu, Yang, and
  Zhou}]{sun2020mobilebert}
Zhiqing Sun, Hongkun Yu, Xiaodan Song, Renjie Liu, Yiming Yang, and Denny Zhou.
  2020.
\newblock \href {https://doi.org/10.18653/v1/2020.acl-main.195}
  {{M}obile{BERT}: a compact task-agnostic {BERT} for resource-limited
  devices}.
\newblock In \emph{Proceedings of the 58th Annual Meeting of the Association
  for Computational Linguistics}, pages 2158--2170, Online. Association for
  Computational Linguistics.

\bibitem[{Tang et~al.(2019)Tang, Lu, Liu, Mou, Vechtomova, and
  Lin}]{tang2019distilling}
Raphael Tang, Yao Lu, Linqing Liu, Lili Mou, Olga Vechtomova, and Jimmy Lin.
  2019.
\newblock \href {http://arxiv.org/abs/1903.12136} {Distilling task-specific
  knowledge from {BERT} into simple neural networks}.
\newblock \emph{CoRR}, abs/1903.12136.

\bibitem[{Tuckute et~al.(2021)Tuckute, Paunov, Kean, Small, Mineroff, Blank,
  and Fedorenko}]{Tuckute2021.05.28.446230}
Greta Tuckute, Alexander Paunov, Hope Kean, Hannah Small, Zachary Mineroff,
  Idan Blank, and Evelina Fedorenko. 2021.
\newblock \href {https://doi.org/10.1101/2021.05.28.446230} {Frontal language
  areas do not emerge in the absence of temporal language areas: A case study
  of an individual born without a left temporal lobe}.
\newblock \emph{bioRxiv}.

\bibitem[{Wang et~al.(2021)Wang, Liu, and Zhang}]{wang2021can}
Cunxiang Wang, Pai Liu, and Yue Zhang. 2021.
\newblock Can generative pre-trained language models serve as knowledge bases
  for closed-book qa?
\newblock \emph{arXiv preprint arXiv:2106.01561}.

\bibitem[{Wang et~al.(2020)Wang, Tang, Duan, Wei, Huang, Cao, Jiang, Zhou
  et~al.}]{wang2020k}
Ruize Wang, Duyu Tang, Nan Duan, Zhongyu Wei, Xuanjing Huang, Cuihong Cao,
  Daxin Jiang, Ming Zhou, et~al. 2020.
\newblock K-adapter: Infusing knowledge into pre-trained models with adapters.
\newblock \emph{arXiv preprint arXiv:2002.01808}.

\bibitem[{Wolf et~al.(2020)Wolf, Debut, Sanh, Chaumond, Delangue, Moi, Cistac,
  Rault, Louf, Funtowicz, Davison, Shleifer, von Platen, Ma, Jernite, Plu, Xu,
  Scao, Gugger, Drame, Lhoest, and Rush}]{wolf-etal-2020-transformers}
Thomas Wolf, Lysandre Debut, Victor Sanh, Julien Chaumond, Clement Delangue,
  Anthony Moi, Pierric Cistac, Tim Rault, Rémi Louf, Morgan Funtowicz, Joe
  Davison, Sam Shleifer, Patrick von Platen, Clara Ma, Yacine Jernite, Julien
  Plu, Canwen Xu, Teven~Le Scao, Sylvain Gugger, Mariama Drame, Quentin Lhoest,
  and Alexander~M. Rush. 2020.
\newblock \href {https://www.aclweb.org/anthology/2020.emnlp-demos.6}
  {Transformers: State-of-the-art natural language processing}.
\newblock In \emph{Proceedings of the 2020 Conference on Empirical Methods in
  Natural Language Processing: System Demonstrations}, pages 38--45, Online.
  Association for Computational Linguistics.

\bibitem[{Xu and Zhao(2021)}]{xu2021dialogue}
Yi~Xu and Hai Zhao. 2021.
\newblock Dialogue-oriented pre-training.
\newblock \emph{arXiv preprint arXiv:2106.00420}.

\bibitem[{Zhao et~al.(2019)Zhao, Wu, Tao, Xu, Zhao, and Yan}]{zhao2019low}
Xueliang Zhao, Wei Wu, Chongyang Tao, Can Xu, Dongyan Zhao, and Rui Yan. 2019.
\newblock Low-resource knowledge-grounded dialogue generation.
\newblock In \emph{International Conference on Learning Representations}.

\bibitem[{Zhao et~al.(2020)Zhao, Wu, Xu, Tao, Zhao, and
  Yan}]{zhao2020knowledge}
Xueliang Zhao, Wei Wu, Can Xu, Chongyang Tao, Dongyan Zhao, and Rui Yan. 2020.
\newblock Knowledge-grounded dialogue generation with pre-trained language
  models.
\newblock In \emph{Proceedings of the 2020 Conference on Empirical Methods in
  Natural Language Processing (EMNLP)}, pages 3377--3390.

\bibitem[{Zhou et~al.(2018)Zhou, Prabhumoye, and Black}]{zhou2018dataset}
Kangyan Zhou, Shrimai Prabhumoye, and Alan~W Black. 2018.
\newblock A dataset for document grounded conversations.
\newblock In \emph{Proceedings of the 2018 Conference on Empirical Methods in
  Natural Language Processing}.

\bibitem[{Zhou et~al.(2021)Zhou, Hedayatnia, Gopalakrishnan, Kim, Pujara, Ren,
  Liu, and Hakkani-Tur}]{zhou2021think}
Pei Zhou, Behnam Hedayatnia, Karthik Gopalakrishnan, Seokhwan Kim, Jay Pujara,
  Xiang Ren, Yang Liu, and Dilek Hakkani-Tur. 2021.
\newblock Think before you speak: Learning to generate implicit knowledge for
  response generation by self-talk.
\newblock In \emph{Proceedings of the 3rd Workshop on Natural Language
  Processing for Conversational AI}, pages 251--253.

\end{thebibliography}
\bibliographystyle{acl_natbib}

\appendix

\clearpage

\section{Additional Cluster Analysis}
\label{sec:cluster_analysis}

We show in Table \ref{tab:topic_keywords} the topic keywords list of each cluster when the pre-defined number of clusters $L=4, 8, 16$ in our Contextualized Topic Model. An additional example for case study is presented in Table~\ref{tab:case_study2}. Similar to the analysis in Section~\ref{sec:case_study}, the provided dialogue history is aligned with the topics of Cluster 3, so the model is able to generate factual correct informative response with solely Expert 3, whereas the other experts are not helpful for the given data sample.

In Figure~\ref{fig:ratio}, we present the ratio of each cluster when $L=4, 8, 16$. From the cluster distribution, we can observe that there is a dominant cluster in the WoW training data across different numbers of clusters. This is because of the nature of the WoW dataset. While setting a larger number of clusters will help the cluster ratio over the training and test sets to be more equal distributed, it will also lead to the problem that there is insufficient training data for each cluster during task adaptation.



\section{Additional Details on Human Evaluation}
\label{sec:human_eval_appd}
We collect human annotations for both humanness and informativeness via crowd-sourcing platform provided by Amazon Mechanical Turk.\footnote{\url{https://www.mturk.com}} For quality control, we limit the annotators' locations to be the United States, United Kingdom, Canada, or Australia to ensure English proficiency. Moreover, we qualify annotators with a HIT Approval rate larger than 95\% and HIT Approved number greater than 5000. As the average time that annotators will spend per response comparison for informativeness is 168 seconds, we reject annotators who spend less than 10 seconds so as to maintain the quality.
The annotator instructions for human evaluation are shown in Figure~\ref{fig:template_human} and Figure~\ref{fig:template_style}. Each annotator is asked to judge either the humanness or informativeness of one dialogue. To get a consistent observation, we use the same 100 randomly selected prefixes of the dialogues across the comparisons. 

\section{Configuration for Inference Efficiency}
We randomly sample 100 data samples from the seen and unseen test set of WoW, respectively. The sampled data are leveraged for all the inference efficiency evaluation experiments. We set the batch size as 1, and repeat each evaluation five times respectively on samples from seen and unseen test set. The final value is the average of ten trials. The device configuration for inference efficiency evaluation is shown in Table~\ref{tab:retrieval-device} and Table~\ref{tab:device}. For the generation inference time evaluation, to have a fair comparison, the generation length is set as 23 for all the models, where 23 is the average response length in the WoW dataset.

\setcounter{figure}{0}
\renewcommand{\thefigure}{A\arabic{figure}}
\begin{figure}[!t]
\centering
\includegraphics[width=0.9\linewidth]{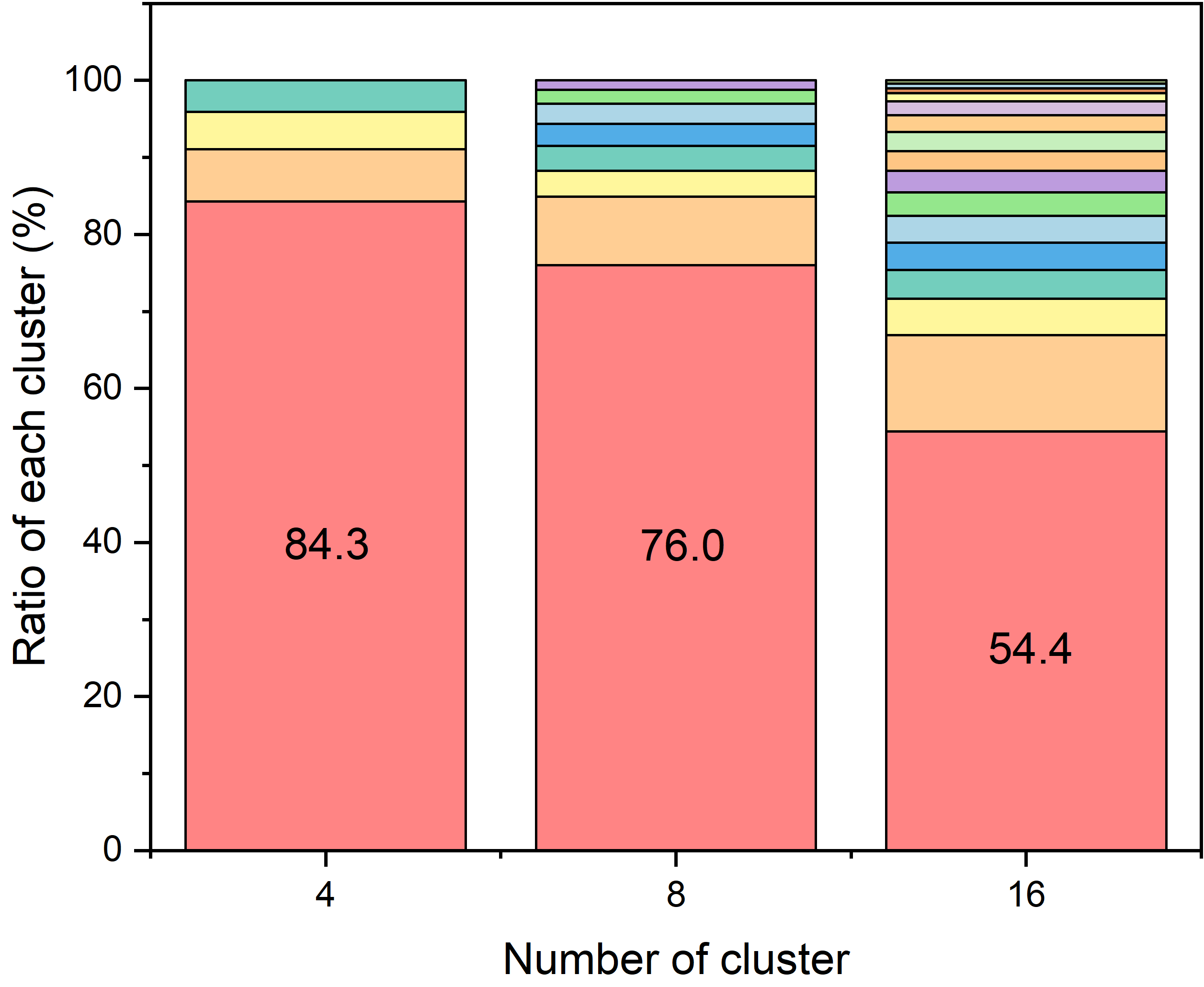} 
\caption{Ratio of the dialogue samples in WoW training set when $L=4,8,16$. From the cluster distribution, we can observe a dominant cluster in the WoW training data.}
\label{fig:ratio}
\end{figure}

\setcounter{table}{0}
\renewcommand{\thetable}{C\arabic{table}}

\begin{table}[h]
\centering
\resizebox{0.95\linewidth}{!}{
    \begin{tabular}{@{}lcc@{}}
    \toprule
    Model  & Device (CPU / GPU)        & \# of Device \\ \midrule
    TF-IDF & Intel Xeon E5-2620 V4 CPU & 1            \\
    DPR    & GeForce GTX 1080Ti        & 1            \\
    GENRE  & GeForce GTX 1080Ti        & 1            \\
    CTM    & Intel Xeon E5-2620 V4 CPU & 1            \\ \bottomrule
    \end{tabular}
    }
    \caption{Device configuration for knowledge retrieval methods and CTM topic modeling.}
     \label{tab:retrieval-device}
\end{table}

\begin{table}[h]
\centering
\resizebox{0.95\linewidth}{!}{
\begin{tabular}{@{}lcc@{}}
\toprule
Model  & Device (CPU / GPU)        & \# of Device \\ \midrule
ZRKGC      & GeForce GTX 1080Ti        & 2            \\
KnowledGPT & GeForce GTX 1080Ti        & 1            \\
Ours       & GeForce GTX 1080Ti        & 1            \\\bottomrule
\end{tabular}
}
\caption{Device configuration for response generation (with knowledge selection if applicable).}
\label{tab:device}
\end{table}

\setcounter{table}{0}
\renewcommand{\thetable}{A\arabic{table}}

\begin{table*}[ht]
\centering
\resizebox{\textwidth}{!}{
\begin{tabular}{ll}
    \toprule
    $L = 4$     &                                                                                                 \\ \midrule
    cluster 1 & east, west, river, south, state, city, area, district, north, center, largest, island, park, states, county             \\
    cluster 2 & rock, band, records, music, team, song, album, club, football, record, league, studio, single, released, professional   \\
    cluster 3 & fiction, story, characters, book, disney, novel, episode, film, films, comic, stars, comics, opera, comedy, character   \\
    cluster 4 & pain, bon, canberra, blocked, rutgers, khalil, edmonton, auckland, auburn, capitals, akron, karim, woodstock, cougars, euro \\ \midrule
    $L=8 $      &                                                                                                                         \\ \midrule
    cluster 1 & systems, theory, data, software, computer, information, person, system, value, mobile, use, users, user, physical, devices \\
    cluster 2 & film, character, characters, episode, comic, television, comedy, fiction, story, comics, directed, novel, films, fictional, fantasy \\
    cluster 3 & company, school, university, students, education, founded, schools, institute, president, department, business, public, united, states, private     \\
    cluster 4 & empire, roman, german, period, chinese, century, russian, soviet, religious, french, bc, king, war, battle, dynasty     \\
    cluster 5 & area, south, city, north, west, ye, located, river, population, east, park, part, county, region, island                \\
    cluster 6 & sugar, yellow, rice, tree, meat, cats, neck, pain, egg, sauce, corn, chicken, breed, hair, cheese                       \\
    cluster 7 & music, band, rock, song, album, records, studio, singer, pop, single, guitar, group, songs, recorded, released          \\
    cluster 8 & league, team, football, club, sports, professional, championship, hockey, baseball, teams, cup, basketball, division, played, racing    \\ \midrule
    $L=16 $      &                                                                                                                      \\ \midrule
    cluster 1 & team, football, league, club, championship, cup, basketball, hockey, wrestling, professional, baseball, olympic, teams, race, rugby \\
    cluster 2 & company, brand, car, chain, ford, owned, cars, corporation, stores, inc, brands, sold, manufacturer, headquartered, restaurant      \\
    cluster 3 & film, episode, directed, stars, fox, drama, comedy, cast, aired, episodes, soap, abc, show, opera, movie                \\
    cluster 4 & light, used, water, surface, temperature, earth, power, energy, materials, chemical, space, speed, material, carbon, electric   \\
    cluster 5 & album, records, song, studio, single, release, track, recorded, lead, songs, chart, recording, label, hot, hit            \\
    cluster 6 & war, army, military, party, navy, ii, forces, election, force, battle, soviet, royal, corps, armed, campaign              \\
    cluster 7 & care, organization, laws, act, tax, organizations, education, non, profit, policy, legal, law, health, rights, agency      \\
    cluster 8 & brain, blood, normal, condition, cause, sleep, eye, causes, fever, heart, psychological, surgery, emotional, loss, drugs    \\
    cluster 9 & ocean, mountain, region, land, coast, pacific, islands, sea, gulf, island, mountains, capital, rivers, km, river            \\
    cluster 10 & computer, digital, data, software, internet, web, code, users, devices, value, mobile, application, device, systems, user  \\
    cluster 11 & street, park, center, road, railway, station, historic, built, building, route, located, highway, opened, city, line      \\
    cluster 12 & century, chinese, greek, christian, modern, medieval, ancient, period, middle, ad, roman, traditions, culture, bc, tradition   \\
    cluster 13 & yellow, bird, tree, flowers, breed, meat, rice, dog, wild, white, sugar, leaf, colour, pepper, flower                      \\
    cluster 14 & professor, father, mother, worked, born, graduated, institute, degree, married, studied, bachelor, moved, mary, graduate, attended \\
    cluster 15 & bass, jazz, guitar, music, festival, stage, dance, theatre, artists, musical, bands, piano, hip, musician, blues      \\
    cluster 16 & fantasy, comics, published, comic, game, fiction, book, universe, books, marvel, created, video, playstation, developed, dc  \\ \bottomrule
    \end{tabular}
    }
    \caption{Top 15 frequent words for each topic cluster of CTM with $L=4, 8, 16$.}
    \label{tab:topic_keywords}
\end{table*}

\begin{table*}[t!]
\centering
\resizebox{0.95\textwidth}{!}{
\begin{tabular}{@{}ll|ll@{}}
\toprule
\multicolumn{2}{l|}{Context} &
  \multicolumn{2}{l}{User: Harry Potter.} \\ \midrule
\multicolumn{1}{l|}{\multirow{8}{*}{\begin{tabular}[c]{@{}l@{}}Case study \\ with single \\ knowledge \\ expert in \\ KnowExpert$_\mathrm{w}$ ($L=4$)\end{tabular}}} &
  \multirow{2}{*}{Expert 1} &
  \multicolumn{1}{l|}{Harry Potter is an American author, investor, philanthropist, and philanthropist.} &
  \multirow{2}{*}{\ding{55}} \\
\multicolumn{1}{l|}{} &
   &
  \multicolumn{1}{l|}{\textit{\textbf{Topics of Cluster 1}: east, river, south, state, city, area, island, ...}} &
   \\ \cmidrule(l){2-4} 
\multicolumn{1}{l|}{} &
  \multirow{2}{*}{Expert 2} &
  \multicolumn{1}{l|}{Harry Potter is an American \underline{musician, songwriter, record producer}, and actor.} &
  \multirow{2}{*}{\ding{55}} \\
\multicolumn{1}{l|}{} &
   &
  \multicolumn{1}{l|}{\textit{\textbf{Topics of Cluster 2}: rock, band, music, album, football, single, ...}} &
   \\ \cmidrule(l){2-4} 
\multicolumn{1}{l|}{} &
  \multirow{2}{*}{Expert 3} &
  \multicolumn{1}{l|}{Harry Potter is a fantasy \underline{novel}, written by J. K. Rowling.} &
  \multirow{2}{*}{\ding{51}} \\
\multicolumn{1}{l|}{} &
   &
  \multicolumn{1}{l|}{\textit{\textbf{Topics of Cluster 3}: fiction, story, characters, novel, film, stars, ...}} &
   \\ \cmidrule(l){2-4} 
\multicolumn{1}{l|}{} &
  \multirow{2}{*}{Expert 4} &
  \multicolumn{1}{l|}{I love Harry Potter, its a great American toy company} &
  \multirow{2}{*}{\ding{55}} \\
\multicolumn{1}{l|}{} &
   &
  \multicolumn{1}{l|}{\textit{\textbf{Topics of Cluster 4}: bon, bucks, rutgers, canberra, ivy, nets, ...}} &
   \\ \midrule
\multicolumn{2}{l|}{KnowExpert$_\mathrm{w}$} &
  \multicolumn{2}{l}{I love Harry Potter.   It’s a great American children’s book series.} \\ \bottomrule
\end{tabular}
}
\caption{Case study on generated responses using the \texttt{KnowExpert$_\mathrm{w}$} model ($L=4$) with the same context on the WoW test unseen set.}
\label{tab:case_study2}
\end{table*}

\setcounter{figure}{0}
\renewcommand{\thefigure}{B\arabic{table}}

\begin{figure*}[t]
\centering
\includegraphics[width=\textwidth]{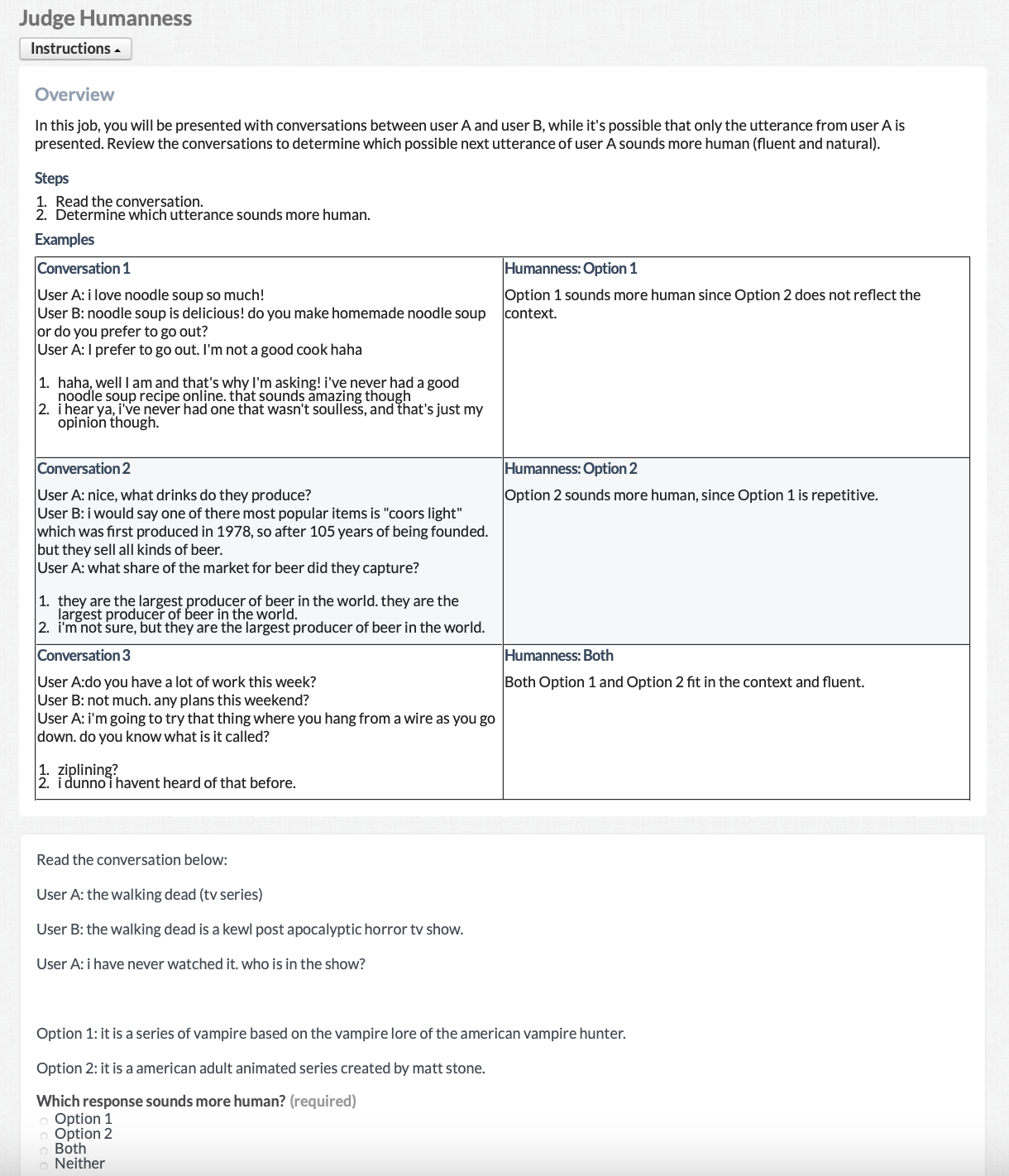}
\caption{Human evaluation template for judging Humanness.}
\label{fig:template_human}
\end{figure*}

\begin{figure*}[t]
\centering
\includegraphics[width=\textwidth]{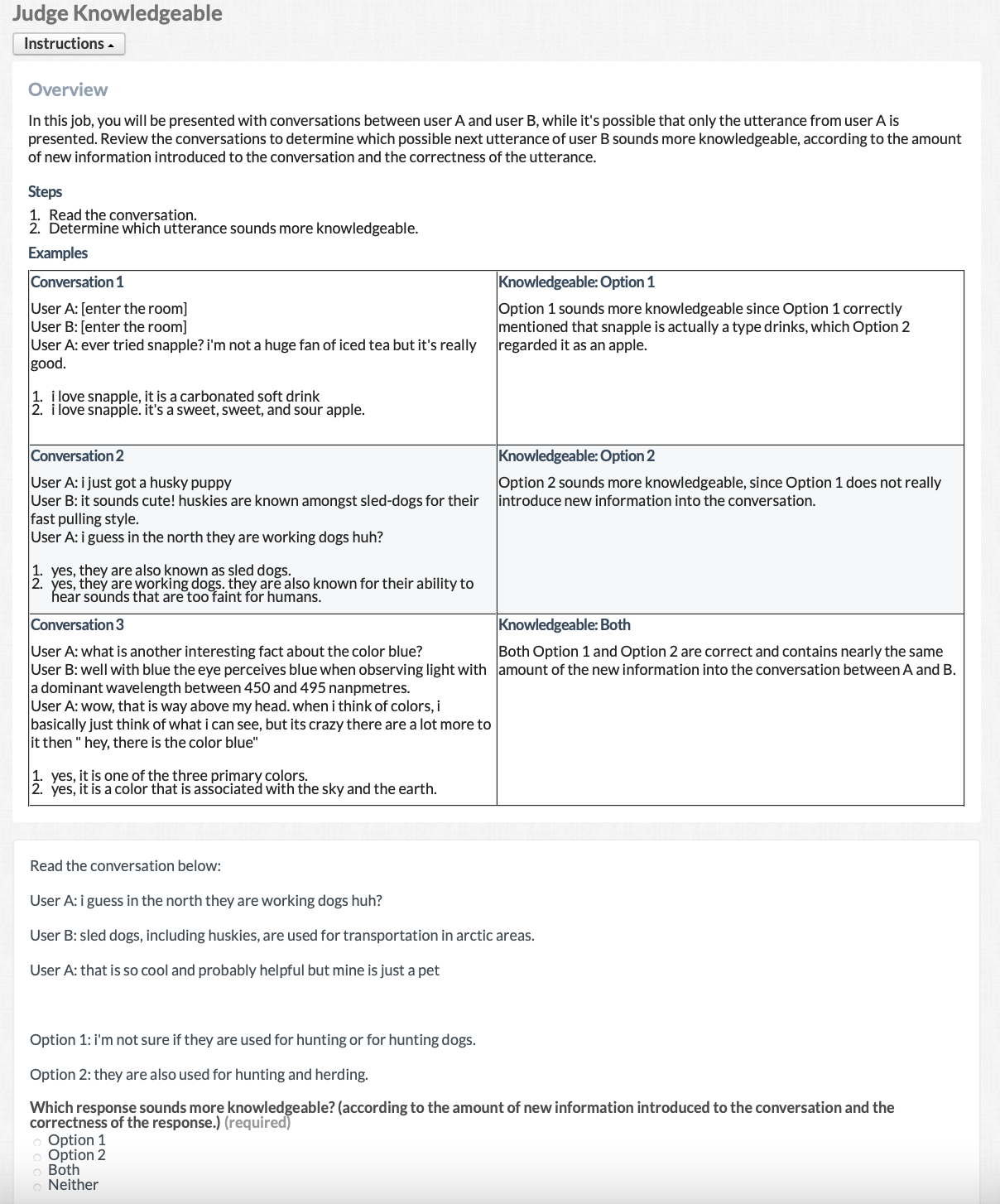}
\caption{Human evaluation template for judging Informativeness.}
\label{fig:template_style}
\end{figure*}

\end{document}